\pdfoutput=1
\documentclass[11pt]{article}

\usepackage[]{ACL2023}

\usepackage{times}
\usepackage{latexsym}
\usepackage{graphicx}

\usepackage[T1]{fontenc}
\usepackage[utf8]{inputenc}

\usepackage{microtype}

\usepackage{inconsolata}

\title{UPB @ ACTI: Detecting Conspiracies using fine tuned Sentence Transformers}

\author{Andrei Paraschiv \and Mihai Dascalu \\
  University Politehnica of Bucharest \\
  313 Splaiul Independetei,  \\
  Bucharest, Romania \\
  \texttt{\{andrei.paraschiv74, mihai.dascalu\}@upb.ro}}

\begin{document}
\maketitle
\begin{abstract}
Conspiracy theories have become a prominent and concerning aspect of online discourse, posing challenges to information integrity and societal trust. As such, we address conspiracy theory detection as proposed by the ACTI @ EVALITA 2023 shared task. The combination of pre-trained sentence Transformer models and data augmentation techniques enabled us to secure first place in the final leaderboard of both sub-tasks. Our methodology attained F1 scores of 85.71\% in the binary classification and 91.23\% for the fine-grained conspiracy topic classification, surpassing other competing systems.
\end{abstract}

\section{Introduction}

Conspiracy theories distort the shared understanding of reality and erode trust in crucial democratic institutions. By substituting reliable, evidence-based information with dubious, implausible, or blatantly false claims, these theories foster a climate of disagreement regarding facts and give undue weight to personal opinions and anecdotal evidence over established facts and scientifically validated theories. \citet{david2010voodoo} defines conspiracy theories as 'the attribution of deliberate agency to something more likely to be accidental or unintended; therefore, it is the unnecessary assumption of conspiracy when other explanations are more probable.' Due to the rapid spread of information across the internet, coupled with the alarming speed at which false information can proliferate \cite{vosoughi2018spread}, we find ourselves amidst what some have dubbed a "golden age" of conspiracy theories \cite{hanley2023golden}. Being a distinct form of misinformation, conspiracy theories exhibit unique characteristics. \citet{brotherton2013measuring} identified five key attributes commonly found in modern conspiracy theories: government malfeasance, extraterrestrial cover-up, malevolent global conspiracies, personal well-being, and information control.

While embracing conspiracy theories can give individuals a sense of reclaiming power or accessing hidden knowledge, these beliefs can sometimes have negative and dangerous consequences. One recent example is the violent insurrection on the US Capitol on 6 January 2021 driven by conspiracy theories surrounding QAnon and election fraud \cite{seitz2021mob}. Additionally, these theories can serve as powerful tools in the hands of nefarious groups, politicians, or state actors who exploit susceptible communities, manipulating them into taking or endorsing actions that can result in significant and dramatic social repercussions \cite{adudeau2023conspiracy, yablokov2022russian}.

Building upon the importance of addressing conspiracy theories, efforts have been made to research and develop automated methods for detecting conspiratorial content on various platforms and languages. For instance, as part of the EVALITA 2023 workshop, the organizers of the ACTI shared task introduced a novel approach: the automatic identification of conspiratorial content in Italian language Telegram messages. This initiative aimed to enhance our ability to quickly recognize and respond to conspiracy theories, enabling the promotion of critical thinking and media literacy by providing reliable sources and encouraging evidence-based discourse. Leveraging such advancements can effectively limit the influence of conspiracy theories while fostering a more informed and resilient society.

This paper presents our contribution to the ACTI @ EVALITA 2023 shared task. We focused on employing the power of pretrained Italian language sentence Transformers. To further enhance the performance and address potential biases, we employed Large Language Models (LLMs) to augment the training data, resulting in a more balanced and comprehensive training set. This combination of leveraging pre-trained models and data augmentation techniques formed the foundation of our methodology, enabling us to achieve first place in the final leaderboard of both sub-tasks with F1 scores of 85.71\% and respectively 91.23\%.    

\section{Related Work}

Recently, online platforms have often banned---entirely deactivated---communities that breached their increasingly comprehensive guidelines.
In 2020 alone, Reddit banned around 2,000 subreddits (the name a community receives on the platform) associated with hate speech.
Similarly, Facebook banned 1,500 pages and groups related to the QAnon conspiracy theory~\cite{qanon2020}. 
While these decisions are met with enthusiasm [e.g., see \citet{adl2020}], the efficacy of ``deplatforming'' these online communities has been questioned~\cite{zuckerman2021deplatforming, russo2023spillover}.
When mainstream platforms ban entire communities for their offensive rhetoric, users often migrate to alternative \emph{fringe platforms}, sometimes created exclusively to host the banned community~\cite{dewey_these_2015, 10.1145/3578503.3583608}.
Banning, in that context, would not only strengthen the infrastructure hosting these fringe platforms~\cite{zuckerman2021deplatforming} but allow these communities to become more toxic elsewhere~\cite{horta2021platform}.
In order to improve the efficacy of such moderation policies identifying and tracking the propagation of problematic content like conspiracy theories is crucial. 
For example the Zika virus outbreak in 2016, coupled with the influence of social networks and the declaration of a public health emergency by the WHO, showed the harm the dissemination of conspiracy theories can generated \cite{ghenai2017zika, wood2018propagating}.

The COVID-19 pandemic had a profound impact, emphasizing the dangers associated with the proliferation of conspiracy theories. These theories encompassed a wide range of topics, including the virus's origin, its spread, the role of 5G networks, and the efficacy and safety of vaccines. With COVID-related lockdowns in place, people became more reliant on social networking platforms such as Twitter, Facebook, and Instagram, which increased their exposure to disinformation and conspiracy theories. MediaEval 2020 \cite{pogorelov2020fakenews} focused on a 5G and COVID-19 conspiracy tweets dataset, proposing two shared tasks to address this issue. The first task involved detecting conspiracies based on textual information, while the second task focused on structure-based detection utilizing the retweet graph. Various systems were proposed to tackle these tasks, employing different approaches such as methods relying on Support Vector Machine (SVM) \cite{moosleitner2020detecting}, BERT \cite{malakhov2020fake}, and GNN \cite{paraschiv2021graph} . In their study, \citet{tyagi2021climate} employed an SVM  to classify the stance of Twitter users towards climate change conspiracies. Their findings revealed that individuals who expressed disbelief in climate change tend to share a significantly higher number of other types of conspiracy-related messages compared to those who believe in climate change. Furthermore, \citet{amin2022detecting} manually labeled 598 Facebook comments as Covid-19 vaccine conspiracy or neutral and used a BERT-based model in conjunction with Google Perspective API to classify these messages, providing valuable insights into the prevalence of vaccine conspiracy theories on social media platforms.

\citet{tunstall2022setfit} presented a new approach based on Sentence Transformers\cite{reimers2019sentence} called SetFit that focused on data-efficient fine-tuning of sentence embeddings, particularly for binary labels. The training of SetFit follows a two-step process. First, it fine-tuned the sentence embeddings in a contrastive manner. This step helped in optimizing the embeddings for the specific classification task. Subsequently, a classification head was trained using finetuned sentence embeddings, enabling effective classification on the training labels. Their approach aimed to enhance the efficiency and performance of fine-tuning sentence embeddings in scenarios with limited data. The efficacy and power of Sentence-Transformers has been shown in multiple tasks spanning from text generation \cite{amin2020exploring, Russo2020ControlGA} to sentence classification tasks.\cite{hong2023empowering, piao2021scholarly, Russo2022DisentanglingAA}. These models capture the semantic and contextual information of sentences or paragraphs, enabling nuanced representations of textual data. Leveraging such models, \citet{bates2023like} used SetFit to propose LAGONN, a hate speech and toxic messages classification framework for content moderation.

\section{Method}
\subsection{Task Description}

The ACTI @ EVALITA 2023 organizers put forth two sub-tasks for participants to address. The first sub-task \cite{acti-subtask-a} involved binary classification, where participants were provided with a dataset consisting of 1,842 training samples and 460 test samples. The objective was to classify messages as either conspiratorial or non-conspiratorial. The second sub-task \cite{acti-subtask-b} focused on fine-grained conspiracy topic classification. Participants were required to classify messages into one of four specific conspiracy topic classes: Covid, QAnon, Flat-Earth, or Russia-conspiracy. A training set of 810 records was provided for this sub-task, while the evaluation test set contained 300 samples. Table \ref{tab:dataset_distribution} shows the class distribution for both sub-tasks. 

\begin{table}[h]\centering
\begin{tabular}{ |p{1.8cm}|l|p{1cm}| }
\hline
 & \textbf{Classes} & \textbf{Count}\\
\hline
Sub-Task A & Non Conspiratorial & 917 \\
 & Conspiratorial & 925 \\
\hline
Sub-Task B & Covid & 435\\
 & QAnon & 242\\
 & Flat-Earth & 76\\
 & Russian & 57\\
\hline
\end{tabular}
\caption{ACTI Dataset distribution for the training sets on Sub-task A and B.}
\label{tab:dataset_distribution}
\end{table}

The macro F1 score was adopted as a criterion to evaluate the two sub-tasks. During the competition, 30\% of the test dataset was immediately evaluated on the Public Leaderboard, giving participants an initial indication of their model's performance. However, the final evaluation was conducted on the remaining 70\% of private entries. These final evaluation scores were then used to compile the Private Leaderboard made public after the conclusion of the competition.
\subsection{Sentence Transformer and Data Augmentation}
\begin{figure}[htp]
  \centering
  \includegraphics[width=0.48\textwidth]{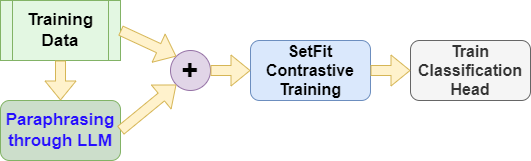}
  \caption{End-to-End training Pipeline.}
  \label{fig:model}
\end{figure}

We considered an Italian language Sentence Transformer model for our submissions and trained contrastive with SetFit\footnote{https://github.com/huggingface/setfit} as described by \citet{tunstall2022setfit}. Since the training dataset is highly imbalanced between the conspiratorial classes (see in Table \ref{tab:dataset_distribution}), we integrated a data augmentation step in our classification pipeline, as seen in Figure \ref{fig:model}.

In the data augmentation step, we used an LLM to create paraphrases for our training data using the prompt "riformulare questo testo: [\textit{comment\_text}]" and different seeds to create variations of the answers. In our experiments, we used "text-davinci-003" from the GPT-3 family\footnote{https://platform.openai.com/docs/models} and the mT5 model finetuned on Italian language paraphrases\footnote{https://huggingface.co/aiknowyou/mt5-base-it-paraphraser}. We set a high temperature (t=0.9) for the LLMs to ensure diverse text generation. The distribution for the augmented dataset is shown in Table \ref{tab:augented_dataset}.

\begin{table}[h]\centering
\begin{tabular}{ |p{1.8cm}|l|p{1cm}| }
\hline
 & \textbf{Classes} & \textbf{Count}\\
\hline
Sub-Task A & Non Conspiratorial & 1,822 \\
 & Conspiratorial & 2,524 \\
\hline
Sub-Task B & Covid & 779\\
 & QAnon & 672\\
 & Flat-Earth & 362\\
 & Russian & 322\\
\hline
\end{tabular}
\caption{Class distribution on the augmented training sets used for Sub-task A and B }
\label{tab:augented_dataset}
\end{table}

\begin{table*}[ht]\centering
\begin{tabular}{ |l|p{2.7cm}| }
\hline
 \textbf{Model} & \textbf{Embedding Size}\\
\hline
    efederici/sentence-BERTino & 768 \\
    efederici/sentence-bert-base & 768 \\
    efederici/sentence-BERTino-3-64 & 64\\
    efederici/mmarco-sentence-BERTino & 768\\
    efederici/sentence-it5-base & 512 \\
    efederici/sentence-it5-small & 512 \\
    nickprock/sentence-bert-base-italian-uncased & 768\\
    nickprock/sentence-bert-base-italian-xxl-uncased & 768\\
    aiknowyou/aiky-sentence-bertino & 768\\
\hline
\end{tabular}
\caption{Sentence-Transformer models considered in our experiments.}
\label{tab:sentence_transformers}
\end{table*}

Sentence-Transformers are pretrained Transformer models finetuned in a Siamese network, such that semantically similar sentences or paragraphs are projected near each other in the embedding space; in contrast, the distance in the embedding space is maximized for sentence pairs that are different. In our experiments, we used several Italian pretrained Sentence Transformers from the Huggingface Hub\footnote{https://huggingface.co/models}, as mentioned in Table \ref{tab:sentence_transformers}. The first step in the SetFit training process involves generating positive and negative triplets. Positive triplets consist of sentences from the same class, while negative triplets contain sentences from different classes. The training data is expanded by including positive and negative triplets, providing a more comprehensive and diverse training set. The Sentence Transformer captures the contextual and semantic information of the messages, providing a powerful feature representation. In the second step, a fully connected classification head is trained on top of the Sentence-Transformer to distinguish between the available classes.

\section{Results}

Besides experimenting with different pre-trained models, as shown in Table \ref{tab:sentence_transformers}, we also performed grid search tuning with several key hyper-parameters, namely the number of iterations, the learning rate, and the number of epochs for training. The number of iterations determined the quantity of generated triplets during training. By adjusting this parameter, we controlled the training data's size, potentially influencing the model's ability to generalize and capture important patterns. We set the maximum sequence length for the tokenizer to 512 for all of our experiments. We withheld 20\% of the training data to evaluate the performance of the trained models during the development time.

The best-performing model differed between the sub-tasks. The best-performing model in the binary classification sub-task was based on "efederici/sentence-BERTino". This model was trained on the "text-davinci-003" augmented dataset for 1 epoch. We used 5 iterations and a learning rate of 1e-05. In contrast, the larger "nickprock/sentence-bert-base-italian-xxl-uncased" model performed best for the fine-grained conspiracy topic classification sub-task. We trained this model on the same dataset for 1 epoch. The learning rate used was 1e-05, and the number of iterations was set to 10. This model yielded the best results in both Leaderboards (see Table \ref{tab:results}).

\begin{table}[h]\centering
\begin{tabular}{ |l|p{2cm}|p{2cm}| }
\hline
  & \textbf{Public Leaderboard} & \textbf{Private Leaderboard}\\
\hline
Sub-Task A & 85.36\% & 85.71\% \\
\hline
Sub-Task B & 87.62\% & 91.23\%\\
\hline
\end{tabular}
\caption{Best model performance on the Public and Private learderboard}
\label{tab:results}
\end{table}

We conducted an ablation study after the competition ended to assess the impact of data augmentation. We trained the best-performing models under different conditions: a) using the original training data with 20\% reserved for development evaluation, b) considering the entire original training data, and c) employing the dataset that was augmented with the mT5 paraphrasing LLM. The results in Tables \ref{tab:results-abl-a} and \ref{tab:results-abl-b} show the importance of the augmentation step. 

\begin{table}[h]\centering
\begin{tabular}{|p{4cm}|p{1.2cm}|p{1.2cm}| }
\hline
  & \textbf{Public Leaderboard} & \textbf{Private Leaderboard}\\
\hline
No augmented data & 75.80\% & 81.29\%\\
\hline
No augmented data with development set & 79.36\% & 83.83\% \\
\hline
mT5 augmented Data & 78.15\% & 82.25\%\\
\hline
\end{tabular}
\caption{Ablation Study for Sub-Task A.}
\label{tab:results-abl-a}
\end{table}

\begin{table}[h]\centering
\begin{tabular}{ |p{4cm}|p{1.2cm}|p{1.2cm}| }
\hline
  & \textbf{Public Leaderboard} & \textbf{Private Leaderboard}\\
\hline
No augmented data & 83.32\% & 93.67\% \\
\hline
No augmented data with development set & 83.39\% & 89.67\%\\
\hline
mT5 augmented Data & 83.24\% & 87.07\%\\
\hline
\end{tabular}
\caption{Ablation Study for Sub-Task B.}
\label{tab:results-abl-b}
\end{table}

In the case of sub-task A, the additional data substantially influenced both the Public and Private test results. The augmented dataset led to significant improvements in performance. However, we see a decline in the Private Leaderboard for the fine-grained task results as the amount of data increased, despite the Public Leaderboard performance keeping the same. This performance decline could be attributed to an unusual distribution difference between the Public and Private test rows. Furthermore, the quality of the paraphrases used in the augmentation process played a crucial role in both sub-tasks. The poor performance achieved by the mT5 model suggests that the quality of the generated paraphrases has a notable impact on the overall model performance. Similarly, a drastic decrease in performance was observed for the second sub-task private leaderboard, arguing for the questionable quality of the paraphrases. 

\section{Conclusion}

In this paper, we described our approach addressing the two sub-tasks in the ACTI @ EVALITA 2023 competition. The challenge focuses on automatically detecting conspiratorial Telegram messages and the classification into four conspiracy topics: Covid, QAnon, Flat-Earth, and Russian conspiracies. Through the utilization of text augmentation techniques and the training of Sentence-Transformers with contrastive learning, we developed robust classifiers. Our best models achieved first place in the Private Leaderboard on both tasks  with F1 scores of 85.712\% in the binary classification and 91.225\% for the fine-grained conspiracy topic classification. This paper contributes to the growing body of research on conspiracy theory detection and emphasizes the effectiveness of leveraging pre-trained models and data augmentation techniques. Our results argue the potential of these approaches in addressing the challenges posed by conspiracy theories and their propagation in online platforms.

\section*{Acknowledgement}

This work was supported by a grant of the Ministry of Research, Innovation, and Digitalization, project CloudPrecis, Contract 344/390020/06.09.2021, MySMIS code: 124812, within POC.

\bibliography{custom}
\bibliographystyle{acl_natbib}

% \appendix

% \section{Example Appendix}
% \label{sec:appendix}

%This is a section in the appendix.

\end{document}